\documentclass[10pt,twocolumn,letterpaper]{article}

\usepackage{cvpr}
\usepackage{times}
\usepackage{epsfig}
\usepackage{graphicx}
\usepackage{amsmath}
\usepackage{amssymb}

\usepackage[ruled]{algorithm2e}
\usepackage{multirow}
\usepackage{array}
\usepackage{ctable} 
\usepackage{mathptmx}
\usepackage{makecell}
\usepackage{arydshln}
\usepackage[capposition=bottom]{floatrow}
\usepackage{comment}
\DeclareMathAlphabet{\mathcal}{OMS}{cmsy}{m}{n}
\usepackage{textcomp}
\usepackage{dblfloatfix}
\usepackage{caption}
\usepackage{subcaption}
\usepackage{url}
\usepackage{textpos}

\newcolumntype{?}[1]{!{\vrule width #1}}

\usepackage[pagebackref=true,breaklinks=true,letterpaper=true,colorlinks=false,bookmarks=false]{hyperref}

\cvprfinalcopy 

\def\cvprPaperID{1517} 

\ifcvprfinal\fi

\begin{document}

\title{2.5D Visual Sound}
\author{Ruohan Gao\thanks{Work done during an internship at Facebook AI Research.}\\
The University of Texas at Austin\\
{\tt\small rhgao@cs.utexas.edu}
\and
Kristen Grauman\\
Facebook AI Research\\
{\tt\small grauman@fb.com}\thanks{On leave from The University of Texas at Austin (\tt{grauman@cs.utexas.edu}).}
}


\maketitle


\begin{abstract}
Binaural audio provides a listener with 3D sound sensation, allowing a rich perceptual experience of the scene.  However, binaural recordings are scarcely available and require nontrivial expertise and equipment to obtain. We propose to convert common monaural audio into binaural audio by leveraging video. The key idea is that visual frames reveal significant spatial cues that, while explicitly lacking in the accompanying single-channel audio, are strongly linked to it. Our multi-modal approach recovers this link from unlabeled video.  We devise a deep convolutional neural network that learns to decode the monaural (single-channel) soundtrack into its binaural counterpart by injecting visual information about object and scene configurations. We call the resulting output \emph{2.5D visual sound}---the visual stream helps  ``lift" the flat single channel audio into spatialized sound. In addition to sound generation, we show the self-supervised representation learned by our network benefits audio-visual source separation. Our video results: \url{http://vision.cs.utexas.edu/projects/2.5D_visual_sound/}
\end{abstract}

\begin{textblock*}{\textwidth}(0cm,-16cm)
\centering
In Proceedings of the IEEE Conference on Computer Vision and Pattern Recognition (CVPR), 2019%
\end{textblock*}


\section{Introduction}
\label{sec:intro}

Multi-modal perception is essential to capture the richness of real-world sensory data and environments. People perceive the world by combining a number of simultaneous sensory streams, among which the visual and audio streams are paramount. In particular, both audio and visual data convey significant \emph{spatial} information.  We see where objects are and how the room is laid out. We also hear them: sound-emitting objects indicate their location, and sound reverberations reveal the room's main surfaces, materials, and dimensions. Similarly, as in the famous cocktail party scenario, while having a conversation at a noisy party, one can hear another voice calling out and turn to face it. The two senses naturally work in concert to interpret spatial signals.

\begin{figure}
	\centering
	\includegraphics[scale=0.44]{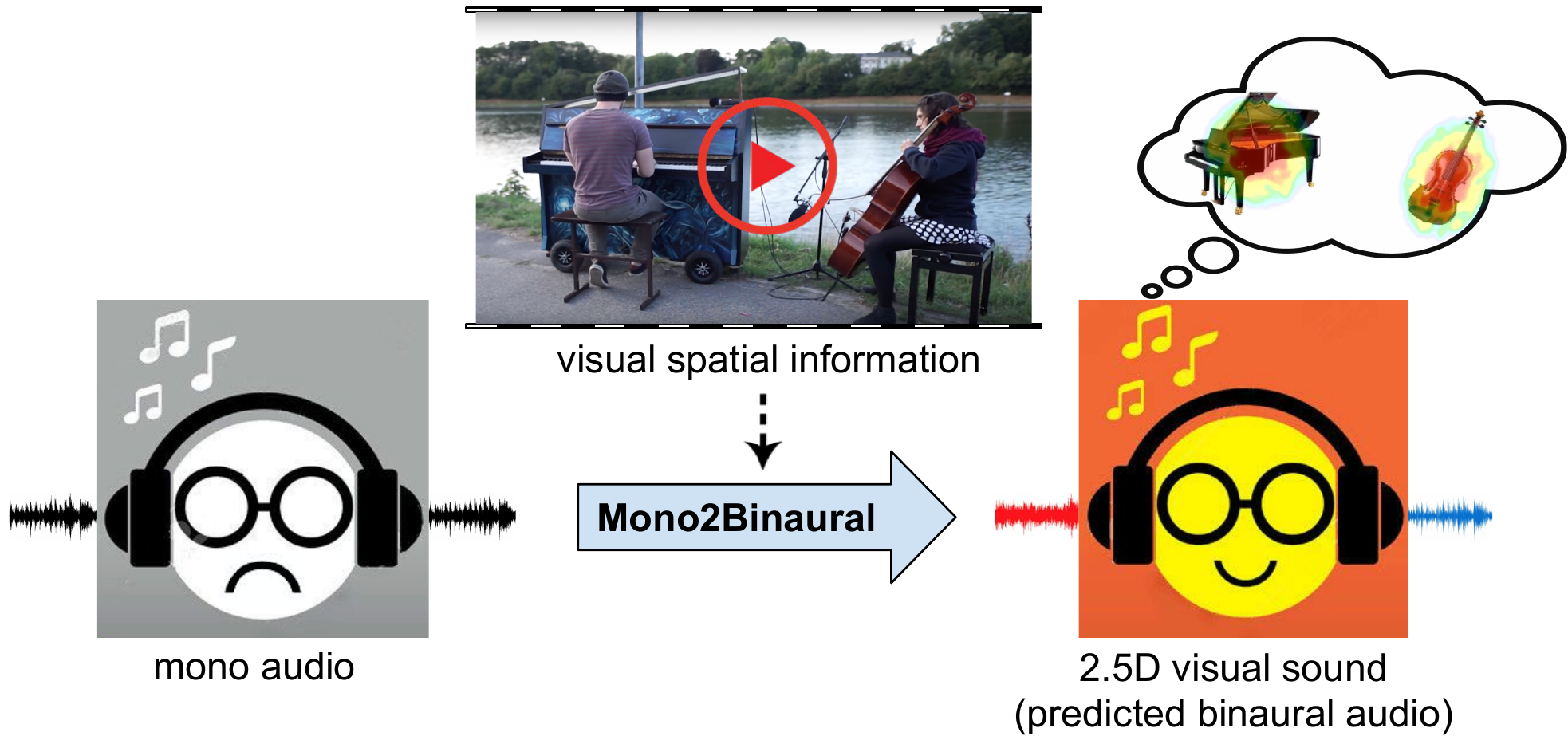}
	\caption{Binaural audio creates a 3D soundscape for listeners, but such recordings remain rare. The proposed approach infers \emph{2.5D visual sound} by injecting the spatial information contained in the video frames accompanying a typical monaural audio stream.}
	\label{fig:concept}
	\vspace*{-0.1in}
\end{figure}

The human auditory system uses \emph{two} ears to extract individual sound sources from a complex mixture. The duplex theory proposed by Lord Rayleigh says that sound source locations are mainly determined by time differences between the sounds reaching each ear (Interaural Time Difference, ITD) and differences in sound level entering the ears (Interaural Level Difference, ILD)~\cite{rayleigh1875our}.
Accordingly, to mimic human hearing, \emph{binaural audio} is usually recorded using two microphones attached to the two ears of a dummy head (see Fig.~\ref{fig:data_collection}).  The rig's two microphones, their spacing, and the physical shape of the ears are all significant for approximating how humans receive sound signals. As a result, when playing binaural audio through headphones, listeners feel the \emph{3D sound sensation} of being in the place where the recording was made and can easily localize the sounds. The immersive spatial sound is valuable for audiophiles, AR/VR applications, and social video sharers alike. 

However, binaural recordings are difficult to obtain in daily life due to the high price of the recording device and the required expertise. Consumer-level cameras typically only record monaural audio with a single microphone, or stereo audio recorded using two microphones with arbitrary arrangement and without physical representation of the pinna (outer ear). We contend that for both machines and people, monaural or even stereo auditory input has very limited dimension. Monaural audio collapses all independent audio streams to the same spatial point, and the listener cannot sense the spatial locations of the sound sources.  

Our key insight is that video accompanying monaural audio has the potential to unlock spatial sound, lifting a flat audio signal into what we call ``2.5D visual sound". Although a single channel audio track alone does not encode any spatial information, its accompanying visual frames do contain object and scene configurations. For example, as shown in Fig.~\ref{fig:concept}, we observe from the video frame that a man is playing the piano on the left and a man is playing the cello on the right. Although we cannot sense the locations of the sound sources by listening to the mono recording, we can nonetheless anticipate what we \emph{would} hear if we were personally in the scene by inference from the visual frames.

We introduce an approach to realize this intuition. Given unlabeled training video, we devise a \textsc{mono2binaural} deep convolutional neural network to convert monaural audio to binaural audio by injecting the spatial cues embedded in the visual frames. Our encoder-decoder style network takes a mixed single-channel audio and its accompanying visual frames as input to perform joint audio-visual analysis, and attempts to predict a two-channel binaural audio that agrees with the spatial configurations in the video. When listening to the predicted binaural audio---the 2.5D visual sound---listeners can then feel the locations of the sound sources as they are displayed in the video. 

Moreover, we show that apart from binaural audio generation, the \textsc{mono2binaural} conversion process can also benefit audio-visual source separation, a key challenge in audio-visual analysis. State-of-the-art systems~\cite{gao2018objectSounds,zhao2018sound,owens2018audio,afouras2018conversation,ephrat2018looking} aim to separate a mixed monaural audio recording into its component sound sources, and thus far they rely solely on the spatial cues evident in the visual stream. We show that the proposed audio-visual binauralization can \emph{self-supervise} representation learning to elicit spatial signals relevant to separation from the audio stream as well. Critically, gaining this new learning signal requires neither semantic annotations nor single-source data preparation, only the same unlabeled binaural training video.

Our main contributions are threefold: Firstly, we propose to convert monaural audio to binaural audio by leveraging video frames, and we design a \textsc{mono2binaural} deep network to achieve that goal; Secondly, we collect FAIR-Play, a $5.2$ hour video dataset with binaural audio---the first dataset of its kind to facilitate research in both the audio and vision communities; Thirdly, we propose to perform audio-visual source separation on predicted binaural audio, and show that it provides a useful self-supervised representation for the separation task. We validate our approach on four challenging datasets spanning a variety of sound sources (e.g.,  instruments, street scenes, travel, sports).

\section{Related Work}

\paragraph{Generating Sounds from Video}
Recent work explores ways to generate audio conditioned on ``silent" video. Material properties are revealed by the sounds objects make when hit with a drumstick, and can be used to synthesize new sounds from silent videos~\cite{owens2016visually}. Recurrent networks~\cite{zhou2017visual} or conditional generative adversarial networks~\cite{chen2017} can generate audio for input video frames, while powerful simulators can synthesize audio-visual data for 3D shapes~\cite{zhang2017gensound}. Rather than generate audio from scratch, our task entails converting an input one-channel audio to two-channel binaural audio guided by the visual frames. 

Only limited prior work considers video-based audio spatialization~\cite{Li2018360audio,morgadoNIPS18}. The system of~\cite{Li2018360audio} synthesizes sound from a speaker in a room as a function of viewing angle, but assumes access to an acoustic impulse recorded in the specific room of interest, which restricts practical use, e.g., for novel ``off-the-shelf" videos. Concurrent work to ours~\cite{morgadoNIPS18} generates \emph{ambisonics} (audio for the full viewing sphere) given 360${}^\circ$ video and its mono audio.  In contrast, we focus on normal field of view (NFOV) video and binaural audio. We show that directly predicting binaural audio  creates better 3D sound sensations for listeners without being restricted to 360${}^\circ$ videos. Moreover, while the end goal of~\cite{morgadoNIPS18} is audio spatialization, we also demonstrate that our \textsc{mono2binaural} conversion process aids audio-visual source separation.

\vspace*{-0.3in}

\paragraph{Audio(-Visual) Source Separation}
Audio-only source separation has been extensively studied in the signal processing literature. ``Blind" separation tackles the case where only a single channel is available~\cite{smaragdis2007supervised,spiertz2009source,virtanen2007monaural,huang2014deep}. Separation becomes easier when multiple channels are observed using multiple microphones~\cite{nakadai2002real,yilmaz2004blind,duong2010under} or binaural audio~\cite{weiss2008source,deleforge2012cocktail,zhang2017deep}. Inspired by this, we transform mono to binaural by observing video, and then leverage the resulting  representation to improve audio-visual separation.

Audio-visual source separation also has a rich history, with methods exploring mutual information~\cite{fisher2001learning}, subspace analysis~\cite{smaragdis2003audio,pu2017audio}, matrix factorization~\cite{parekh2017motion,sedighin2016two,gao2018objectSounds}, and correlated onsets~\cite{barzelay2007harmony,li2017see}.  Recent methods leverage deep learning for audio-visual separation of speech~\cite{ephrat2018looking,owens2018audio,afouras2018conversation,gabbay2017visual}, musical instruments~\cite{zhao2018sound}, and other objects~\cite{gao2018objectSounds}. New tasks are also emerging, such as learning to separate on- and off-screen sounds~\cite{owens2018audio}, learning object sound models from unlabeled video~\cite{gao2018objectSounds}, or predicting sounds per pixel~\cite{zhao2018sound}. All these methods exploit mono audio cues to perform audio-visual source separation, whereas we propose to predict binaural cues to enhance separation. Furthermore, different from the task of localizing pixels responsible for a given sound~\cite{kidron2005pixels,hershey2000audio,zunino,arandjelovic2017objects,zhao2018sound,Senocak_2018_CVPR,tian2018audio}, our goal is to perform binaural audio synthesis.

\paragraph{Self-Supervised Learning}
Self-supervised learning exploits labels freely available in the structure of the data, and audio-visual data offers a wealth of such tasks. Recent work explores self-supervision for visual~\cite{owens2016ambient,arandjelovic2017look} and audio~\cite{aytar2016soundnet} feature learning, cross-modal representations~\cite{aytar2017see}, and audio-visual alignment~\cite{owens2018audio,Korbar2018cotraining,harwath2018jointly}. Our \textsc{mono2binaural} formulation is also self-supervised, but unlike any of the above, we use visual frames to supervise audio spatialization, while also learning better sound representations for audio-visual source separation.

\section{Approach}

Our approach learns to map monaural audio to binaural audio via video.  In the following, we first describe our binaural audio video dataset (Sec.~\ref{sec:data}).  Then we present our \textsc{mono2binaural} formulation (Sec.~\ref{sec:formulate}), and our network and training procedure to solve it (Sec.~\ref{sec:network}).  Finally we introduce our approach to leverage inferred binaural sound to perform audio-visual source separation (Sec.~\ref{sec:separate}).

\subsection{FAIR-Play Data Collection}\label{sec:data}
\begin{figure}
	\centering
	\includegraphics[scale=0.635]{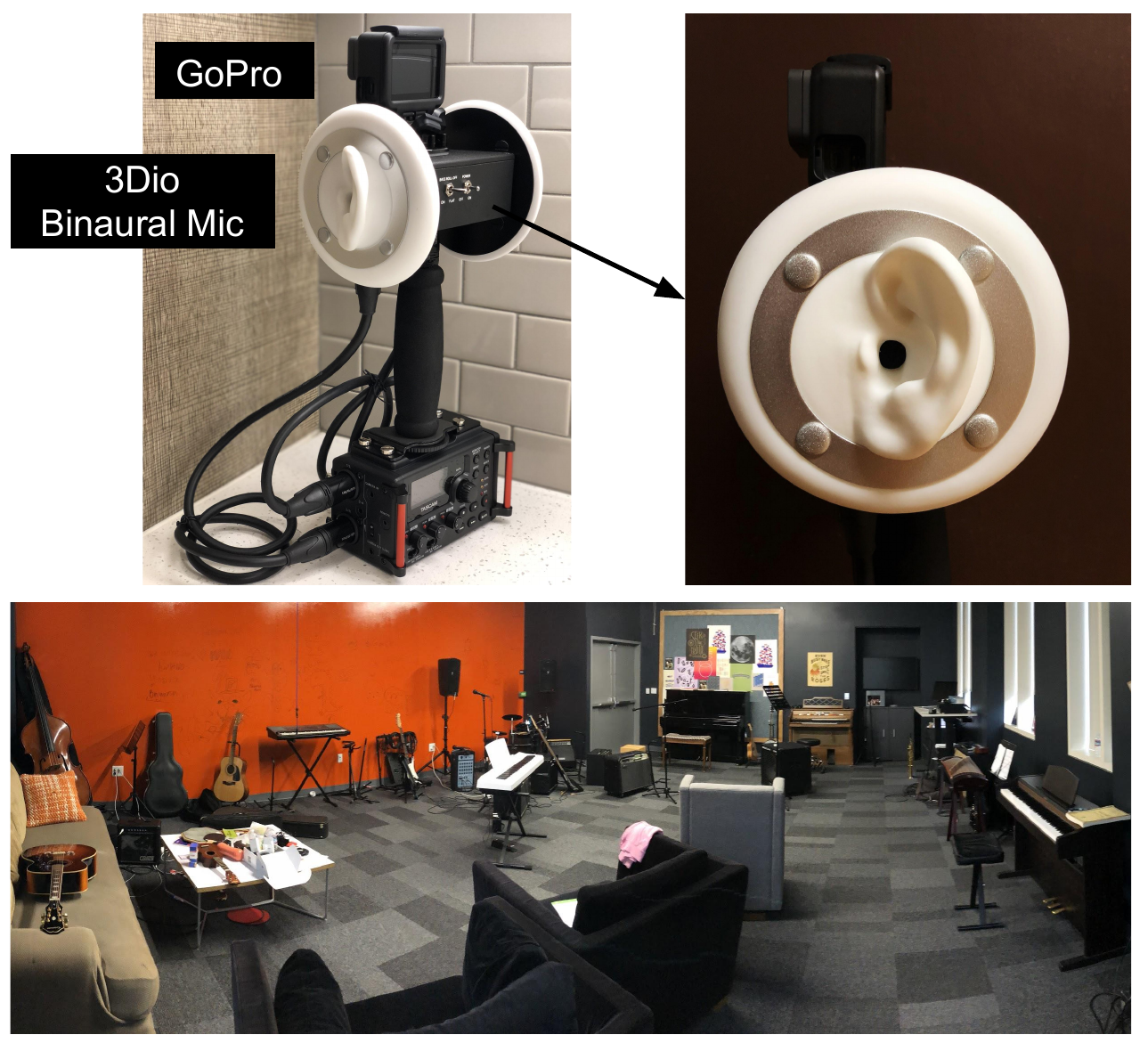}
	\caption{Binaural rig and data collection in a music room.}
	\label{fig:data_collection}
	\vspace*{-0.1in}
\end{figure}

Training our method requires binaural audio and accompanying video. Since  no large public video datasets contain binaural audio, we collect a new dataset we call FAIR-Play with a custom rig. As shown in Fig.~\ref{fig:data_collection}, we assembled a rig consisting of a 3Dio Free Space XLR binaural microphone, a GoPro HERO6 Black camera, and a Tascam DR-60D recorder as the audio pre-amplifier. We mounted the GoPro camera on top of the 3Dio binaural microphone to mimic a person's embodiment for seeing and hearing, respectively. The 3Dio binaural microphone records binaural audio, and the GoPro camera records videos at 30fps with stereo audio. We simultaneously record from both devices so the streams are roughly aligned.

Note that both the ear shaped housing (pinnae) for the microphones and their spatial separation are significant; professional binaural mics like 3Dio simulate the physical manner in which humans receive sound. In contrast, stereo sound is captured by two mics with an arbitrary separation that varies across capture devices (phones, cameras),  and so lacks the spatial nuances of binaural.  The limit of  binaural capture, however, is that a single rig inherently assumes a single \emph{head-related transfer function}, whereas individuals have slight variations due to inter-person anatomical differences. Personalizing head-related transfer functions is an area of active research~\cite{iida2014personalization,torres2015personalization}.

We captured videos with our custom rig in a large music room (about 1,000 square feet). Our intent was to capture a variety of sound making objects in a variety of spatial contexts, by assembling different combinations of instruments and people in the room. The room contains various instruments including cello, guitar, drum, ukelele, harp, piano, trumpet, upright bass, and banjo. We recruited 20 volunteers to play and recorded them in solo, duet, and multi-player performances. We post-process the raw data into 10s clips. In the end, our FAIR-Play\footnote{\label{dataset_page}\url{https://github.com/facebookresearch/FAIR-Play}} dataset consists of 1,871 short clips of musical performances, totaling 5.2 hours. In experiments we use both the music data as well as ambisonics datasets~\cite{morgadoNIPS18} for street scenes and YouTube videos of sports, travel, etc. (cf.~Sec.~\ref{sec:results}).

\subsection{Mono2Binaural Formulation}\label{sec:formulate}
Binaural cues let us infer the location of sound sources. The interaural time difference (ITD) and the interaural level difference (ILD)  play an essential role. ITD is caused by the difference in travel distances between the two ears. When a sound source is  closer to one ear than the other, there is a time delay between the signals' arrival at the two ears. ILD is caused by a ``shadowing" effect---a listener's head is large relative to certain wavelengths of sound, so it serves as a barrier, creating a shadow. The particular shape of the head, pinnae, and torso also act as a filter depending on the locations of the sound sources (distance, azimuth, and elevation). All these cues are missing in monaural audio, thus we cannot sense any spatial effect by listening to single-channel audio.

\begin{figure*}[t]
    \center
    \includegraphics[scale=0.72]{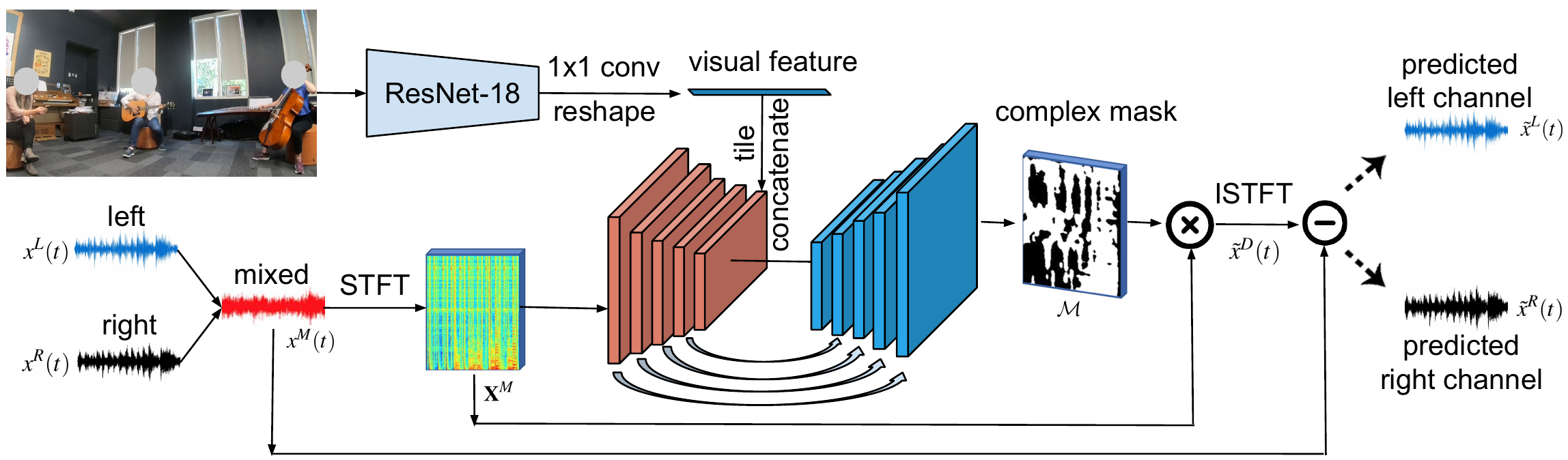}
    \caption{Our \textsc{mono2binaural} deep network takes a mixed monaural audio and its accompanying visual frame as input, and predicts a two-channel binaural audio output that satisfies the visual spatial configurations. An ImageNet pre-trained ResNet-18 network is used to extract visual features, and a U-NET is used to extract audio features and perform joint audio-visual analysis. We predict a complex mask for the audio difference signal, then combine it with the input mono audio to restore the left and right channels, respectively. At test time, the input is single-channel monaural audio.}
    \label{fig:network}
    \vspace*{-0.1in}
\end{figure*}

We denote the signal received at the left and right ears by $x^L(t)$ and $x^R(t)$, respectively.  If we mix the two channels into a single channel $x^M(t) = x^L(t) + x^R(t)$, then all spatial information collapses. We can formulate a self-supervised task to take the mixed monaural signal $x^M(t)$ as input and split it into two separate channels $\tilde{x}^L(t)$ and $\tilde{x}^R(t)$, using the original 
$x^L(t)$, $x^R(t)$ as ground-truth during training. However, this is a highly under-constrained problem, as $x^M(t)$ lacks the necessary information to recover both channels.   Our key idea is to guide the \textsc{mono2binaural} process with the accompanying video frames, from which \emph{visual} spatial information can serve as supervision.

Instead of directly predicting the two channels, we predict the difference of the two channels:
\begin{equation}
	\vspace*{-0.05in}
x^D(t) = x^L(t) - x^R(t).
	\vspace*{-0.05in}
\end{equation}

More specifically, we operate on the frequency domain and perform short-time Fourier transform (STFT)~\cite{griffin1984signal} on $x^M(t)$ to obtain the complex-valued spectrogram $\textbf{X}^M$, and  the objective is to predict the complex-valued spectrogram $\textbf{X}^D$ for $x^D(t)$:
\begin{equation}
	\textbf{X}^M = \{ \textbf{X}^{M}_{t,f} \}_{t=1,f=1}^{T,F}, ~~~~~\textbf{X}^D = \{ \textbf{X}^{D}_{t,f} \}_{t=1,f=1}^{T,F},
\end{equation}
where $t$ and $f$ are the time frame and frequency bin indices, respectively, and $T$ and $F$ are the numbers of bins. Then we obtain the predicted difference signal $\tilde{x}^D(t)$ by the inverse short-time Fourier transform (ISTFT)~\cite{griffin1984signal} of $\textbf{X}^{D}$. Finally, we recover both channels---the binaural audio output:
\begin{equation}
\tilde{x}^L(t) =  \frac{x^M(t) + \tilde{x}^D(t)}{2},~~~~~\tilde{x}^R(t) =  \frac{x^M(t) - \tilde{x}^D(t)}{2}.\label{eq:channels}
\end{equation}

\subsection{Mono2Binaural Network}\label{sec:network}
\label{sec:network}
Next we present our \textsc{mono2binaural} deep network to perform audio spatialization. The network takes the mono audio $x^M(t)$ and visual frames as input and predicts $x^D(t)$.  

As shown in Fig.~\ref{fig:network}, we extract visual features from the center frame of the audio segment using ResNet-18~\cite{he2016deep}, which is pre-trained on ImageNet. The ResNet-18 network extracts per-frame features after the $4^{th}$ ResNet block with size $(H/32) \times (W/32) \times C$, where $H,W,C$ denote the frame and channel dimensions. We then pass the visual feature through a $1 \times 1$ convolution layer to reduce the channel dimension, and flatten it into a single visual feature vector.

On the audio side, we adopt a U-NET~\cite{ronneberger2015u} style architecture. The U-NET encoder-decoder network adopted here is ideal for our dense prediction task where the input and output have the same dimension. We mix the left and right channels of the binaural audio, and extract a sequence of STFT frames to generate an audio spectrogram $\textbf{X}^M$. We use the complex spectrogram: each time-frequency bin contains the real and imaginary part of the corresponding complex spectrogram value. Then it is passed through a series of convolution layers to extract an audio feature of dimension $ (T/32) \times (F/32) \times C$. We replicate the visual feature vector $(T/32) \times (F/32)$ times, tile them to match the audio feature dimension, and then concatenate the audio and visual feature maps along the channel dimension. Through the series of operations, each audio feature dimension is injected with the visual feature to perform joint audio-visual analysis.

Finally, we perform up-convolutions on the concatenated audio-visual feature map to generate a complex multiplicative spectrogram mask $\mathcal{M}$. In source separation tasks, spectrogram masks have proven better than alternatives such as direct prediction of spectrograms or raw waveforms~\cite{wang2018supervised}. Similarly, here we also adopt the idea of masking, but our goal is to mask the spectrogram of the mixed mono audio and predict the spectrogram of the difference signal, rather than perform separation. The real and imaginary components of the complex mask are separately estimated in the real domain. We add a sigmoid layer after the up-convolution layers to bound the complex mask values to [-1, 1], similar to~\cite{ephrat2018looking}. The series of convolutions and up-convolutions maps the input mono spectrogram to a complex mask that encodes the predicted binaural audio.

Initially, we attempted to directly predict the left and right channels. However, we found that direct prediction makes the network fall back on a ``safe" but useless solution of copying and pasting the input audio, without reasoning with the visual features. Instead, predicting the difference signal forces the deep network to analyze the visual information and learn the subtle difference between the two channels, as required by the binaural audio target.

The spectrogram of the difference signal is then obtained by complex multiplying the input spectrogram with the predicted complex mask: 
\begin{equation}
	\vspace*{-0.075in}
    \tilde{\textbf{X}}^D = \mathcal{M} \cdot \textbf{X}^M.
    \vspace*{-0.015in}
\end{equation}
We train our \textsc{mono2binaural} network using L2 loss to minimize the distance between the ground-truth complex spectrogram and the predicted one. Finally, using ISTFT, we obtain the predicted difference signal $\tilde{x}^D(t)$, through which we recover the two channels $\tilde{x}^L(t)$ and $\tilde{x}^R(t)$ as defined in Eq.~\ref{eq:channels}. See supp. for network details.

At test time, the network is presented with monaural audio and a video frame and infers the binaural output, i.e., the 2.5D visual sound. To process a full video stream, each video is decomposed into many short audio segments.  Video frames usually do not change much within such a short segment. We use a sliding window to perform spatialization segment by segment with a small hop size, and average predictions on overlapping parts. Thus, our method is able to handle moving sound sources and cameras.

Our approach expects a similar field of view (FoV) between training and testing, and assumes the microphone is near the camera.  Our experiments demonstrate we can learn \textsc{mono2binaural} for both normal FoV and 360${}^\circ$ video, and furthermore the same system can cope with mono inputs from variable hardware (e.g., YouTube videos).

\subsection{Audio-Visual Source Separation}\label{sec:separate}

So far we have defined our \textsc{mono2binaural} approach to convert monaural audio to binaural audio by introducing visual spatial cues from video.  Recall that we have two goals: to predict binaural audio for sound generation itself, and to explore its utility for audio-visual source separation. 

Audio source separation is the problem of obtaining an estimate for each of the $J$ sources $s_j$ from the observed linear mixture $x(t) = \sum_{j=1}^{J}s_j(t)$. For binaural audio source separation, the problem is to obtain an estimate for each of the $J$ sources $s_j$ from the observed binaural mixture $x^L(t)$ and $x^R(t)$:
\vspace*{-0.15in}
\begin{equation}
	\vspace*{-0.05in}
	x^L(t) = \sum_{j=1}^{J}s^L_j(t), ~~~~~~ x^R(t) = \sum_{j=1}^{J}s^R_j(t),
	\vspace*{-0.05in}
\end{equation}
where $s^L_j(t)$ and $s^R_j(t)$ are time-discrete signals received at the left ear and the right ear for each source, respectively.  

Interfering sound sources are often located at different spatial positions in the physical space.  Human listeners exploit the spatial information from the coordination of both ears to resolve sound ambiguity caused by multiple sources.  This ability is greatly diminished when listening with only one ear, especially in reverberant environments~\cite{koenig1950subjective}. Audio source separation by machine listeners is similarly handicapped, typically lacking access to binaural audio~\cite{zhao2018sound,gao2018objectSounds,owens2018audio,ephrat2018looking}. However, we hypothesize that our \textsc{mono2binaural} \emph{predicted} binaural audio can aid separation.  Intuitively, by forcing the network to learn how to lift mono audio to binaural, its representation is encouraged to expose the very spatial cues that are valuable for source separation. Thus, even though the \textsc{mono2binaural} features see the same video as any other audio-visual separation method, they may better decode the latent spatial cues because of their binauralization ``pre-training" task.

\begin{figure}[t]
    \center
    \includegraphics[scale=0.43]{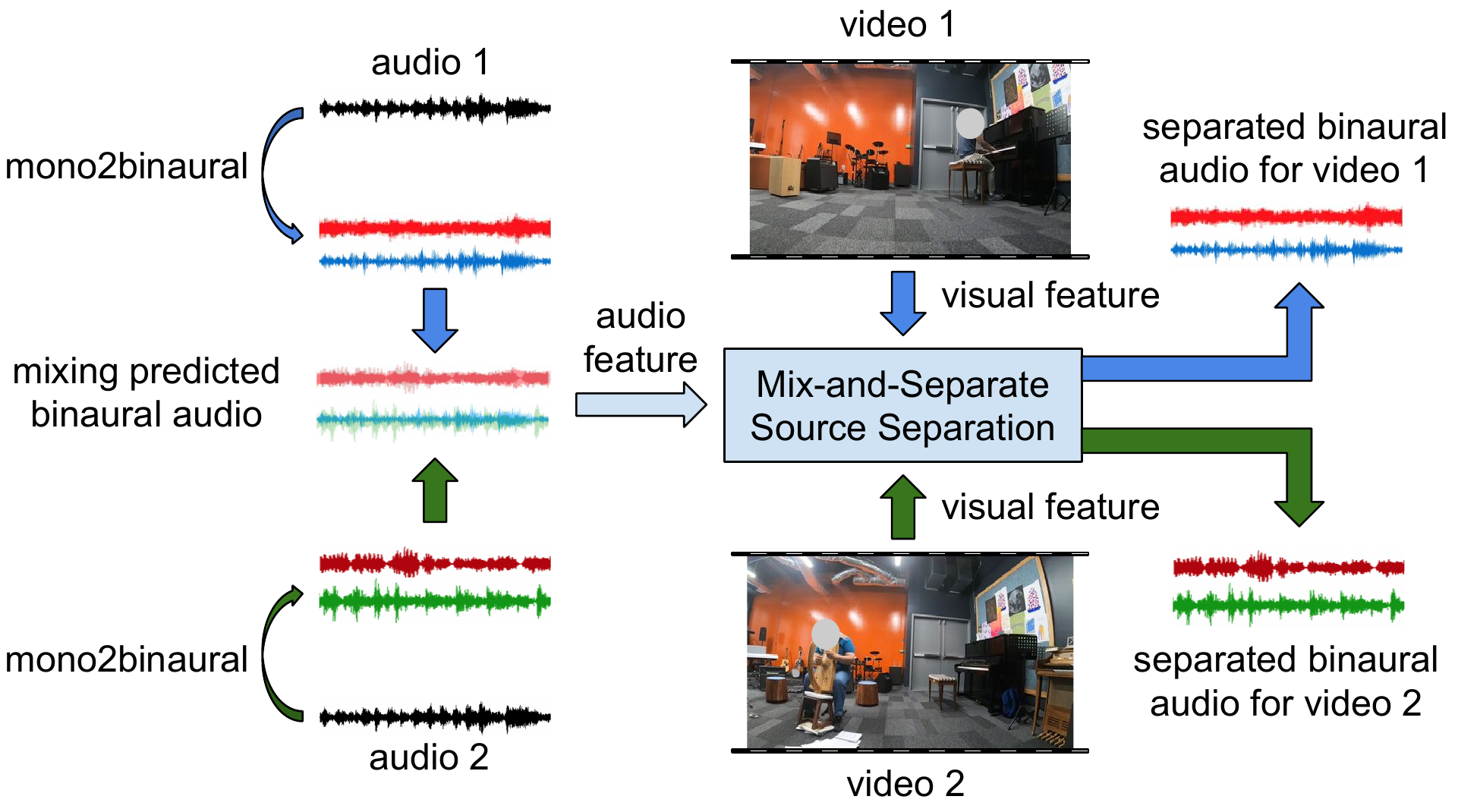}
    \caption{Mix-and-Separate~\cite{zhao2018sound,owens2018audio,ephrat2018looking}-inspired framework for audio-visual source separation. During training, we mix the binaural audio tracks for a pair of videos to generate a mixed audio input. The network learns to separate the sound for each video conditioned on their visual frames.}
    \label{fig:source_separation}
    \vspace*{-0.12in}
\end{figure}

In particular, we expect two main effects. First, binaural audio embeds information about the spatial distribution of sound sources, which can act as a regularizer for separation. Second, binaural cues may be especially helpful in cases where sound sources have similar acoustic characteristics, since the spatial organization can reduce source ambiguities. Related regularization effects are observed in other vision tasks. For example, hallucinating motion enhances static-image action recognition~\cite{gao2018im2flow}, or predicting semantic segmentation informs depth estimation~\cite{liu2010single}.

To implement a testbed for audio-visual source separation, we adopt the Mix-and-Separate idea~\cite{zhao2018sound,owens2018audio,ephrat2018looking}. We use the same base architecture as our \textsc{mono2binaural} network except that now the input to the network is a pair of training video clips. Fig.~\ref{fig:source_separation} illustrates the separation framework. We mix the sounds of the predicted binaural audio for the two videos to generate a complex audio input signal, and the learning objective is to separate the binaural audio for each video conditioned on their corresponding visual frames. Following~\cite{zhao2018sound}, we only use spectrogram magnitude and predict a ratio mask for separation. Per-pixel L1 loss is used for training. See supp. for details.

\section{Experiments}\label{sec:results}

We validate our approach for generation and separation.

\subsection{Datasets}
We use four challenging datasets spanning a wide variety of sound sources, including musical instruments, street scenes, travel, and sports.

\vspace{-0.15in}

\paragraph{FAIR-Play}
Our new dataset consists of 1,871 10s clips of videos recorded in a music room (Fig.~\ref{fig:data_collection}). The videos are paired with binaural audios of high quality recorded by a professional binaural microphone. We create 10 random splits by splitting the data into train/val/test splits of 1,497/187/187 clips, respectively.

\vspace{-0.15in}

\paragraph{REC-STREET}
A dataset collected by~\cite{morgadoNIPS18} using a Theta V 360${}^\circ$ camera with TA-1 spatial audio microphone. It consists of 43 videos (3.5 hours) of  outdoor street scenes.

\vspace{-0.15in}

\paragraph{YT-CLEAN}
This dataset contains in-the-wild 360${}^\circ$ videos from YouTube crawled by~\cite{morgadoNIPS18} using queries related to spatial audio. It consists of 496 videos of a small number of super-imposed sources, such as people talking in a meeting room, outdoor sports, \etc. 

\vspace{-0.15in}

\paragraph{YT-MUSIC}
A dataset that consists of 397 YouTube videos of music performances collected by~\cite{morgadoNIPS18}. It is their most challenging dataset due to the large number of mixed sources (voices and instruments).

To our knowledge, FAIR-Play is the first dataset of its kind that contains videos of professional recorded binaural audio. For REC-STREET, YT-CLEAN and YT-MUSIC, we split the videos into 10s clips and divide them into train/val/test splits based on the provided split1. These datasets only contain ambisonics, so we use a binaural decoder to convert them to binaural audio. Specifically, we use the head related transfer function (HRTF) from NH2 subject in the ARI HRTF Dataset\footnote{\url{http://www.kfs.oeaw.ac.at/hrtf}} to perform decoding. For our FAIR-Play dataset, half of the training data is used to train the \textsc{mono2binaural} network, and the other half is reserved for audio-visual source separation experiments.

\subsection{Implementation Details}\label{exp:implementaiton}
Both our \textsc{mono2binaural} and separation networks are in PyTorch. For all experiments, we resample the audio at 16kHz and STFT is computed using a Hann window of length 25ms, hop length of 10ms, and FFT size of 512. For \textsc{mono2binaural} training, we randomly sample audio segments of length 0.63s from each 10s audio clip. During testing, we use a sliding window with hop size 0.05s to binauralize 10s audio clips for both our method and baselines. For source separation experiments, we use similar network design and training/testing strategies. See supp. for details.

\subsection{Mono2Binaural Generation Accuracy}~\label{exp:mono2binaural}
We evaluate the quality of our predicted binaural audio by using common metrics as well as two user studies. We compare to the following baselines:

\vspace{-0.1in}
\begin{itemize}
\itemsep0em
\item \textbf{Ambisonics~\cite{morgadoNIPS18}}: We use the pre-trained models provided by~\cite{morgadoNIPS18} to predict ambisonics. The models are trained on the same data as our method. Then we use the binaural decoder to convert the predicted ambisonics to binaural audio. This baseline is not available for the BINAURAL-MUSIC-ROOM dataset.

\item \textbf{Audio-Only:} To determine if visual information is essential to perform \textsc{mono2binaural} conversion, we remove the visual stream and implement a baseline using only audio as input. All other settings are the same except that only audio features are passed to the up-convolution layers for binaural audio prediction.

\item \textbf{Flipped-Visual:} During testing, we flip the accompanying visual frames of the mono audios to perform prediction using the wrong visual information. 

\item \textbf{Mono-Mono:} A straightforward baseline that copies the mixed monaural audio onto both channels to create a fake binaural audio.
\end{itemize}
\vspace{-0.1in}

\begin{table*}
\begin{tabular}{c?{0.5mm}cc?{0.5mm}cc?{0.5mm}cc?{0.5mm}cc}

\multirow{2}{*}{}                                     & \multicolumn{2}{c?{0.5mm}}{FAIR-Play} & \multicolumn{2}{c?{0.5mm}}{REC-STREET} & \multicolumn{2}{c?{0.5mm}}{YT-CLEAN}  & \multicolumn{2}{c}{YT-MUSIC}  \\ \cline{2-9} 
                                                      & ~~~~STFT~~      & ~~ENV~~      & STFT  & ENV  & STFT  & ENV
                                                      & STFT  & ENV \\ \specialrule{.12em}{.1em}{.1em}

Ambisonics~\cite{morgadoNIPS18} &        ~~~-          &         -              &      0.744       &    0.126        & 1.435 &  0.155     &    1.885         &    0.183               \\ 
Audio-Only                                            &        ~~~0.966          &       0.141                &     0.590        &      0.114             &    1.065  &  0.131 &  1.553       &      0.167             \\ 
Flipped-Visual                                        &      ~~~1.145            &           0.149            &     0.658        &      0.123             & 1.095  & 0.132 &  1.590        &       0.165            \\ 
Mono-Mono                                             &     ~~~1.155             &         0.153              &     0.774        &      0.136             &  1.369  &  0.153 & 1.853         &           0.184        \\ 
\textsc{Mono2Binaural} (Ours)                                  &     ~~~\textbf{0.836}             &          \textbf{0.132}             &     \textbf{0.565}        &      \textbf{0.109}      & \textbf{1.027} &   \textbf{0.130}    &      \textbf{1.451}       &    \textbf{0.156}               \\ \specialrule{.12em}{.1em}{.1em}

\end{tabular}
\caption{Quantitative results of binaural audio prediction on four diverse datasets. We report the STFT distance and the envelope distance; lower is better. For FAIR-Play, we report the average results across 10 random splits. The results have a standard error of approximately $5 \times 10^{-2}$ for STFT distance and $3 \times 10^{-3}$ for ENV distance on average.}
\label{tab:mono2binaural}
\vspace{-0.1in}
\end{table*}

We report two metrics: 1) \textbf{STFT Distance:} The euclidean distance between the ground-truth and predicted complex spectrograms of the left and right channels:
\vspace*{-0.1in}
$$\mathcal{D}_{\{\text{STFT}\}} = ||\textbf{X}^L - \tilde{\textbf{X}}^L||_2 + ||\textbf{X}^R - \tilde{\textbf{X}}^R||_2.$$ 
2) \textbf{Envelope (ENV) Distance:} Direct comparison of raw waveforms may not capture perceptual similarity well. Following~\cite{morgadoNIPS18}, we take the envelope of the signals, and measure the euclidean distance between the envelopes of the ground-truth left and right channels and the predicted signals. Let $E[x(t)]$ denote the envelope of signal $x(t)$.  The envelope distance is defined as:
\vspace{-0.05in}
$$\mathcal{D}_{\{\text{ENV}\}} = ||E[x^L(t)] - E[\tilde{x}^L(t)||_2 + ||E[x^R(t)] - E[\tilde{x}^R(t)||_2.$$
\vspace{-0.15in}

\vspace*{-0.15in}
\paragraph{Results.}
Table~\ref{tab:mono2binaural} shows the binaural generation results. Our method outperforms all baselines consistently on all four datasets. Our \textsc{mono2binaural} approach performs better than the Audio-Only baseline, indicating the visual stream is essential to guide conversion. Note that the Audio-Only baseline uses the same network design as our method, so it has reasonably good performance.  Still, we find our method outperforms it most when object(s) are not simply located in the center. Flipped-Visual performs much worse, demonstrating that our network properly learns to localize sound sources to predict binaural audio correctly. 

The Ambisonics~\cite{morgadoNIPS18} approach does not do as well. We hypothesize several reasons. The method predicts four channel ambisonics directly, which must be converted to binaural audio. While ambisonics have the advantage of being a more general audio representation that is ideal for 360${}^\circ$ video, predicting ambisonics first and then decoding to binaural audio for deployment can introduce artifacts that make the binaural audio less realistic. Better head-related transfer functions could help to render more realistic binaural audio from ambisonics, but this remains active research~\cite{noisternig20033d,kronlachner2014spatial}.\footnote{We experimented with multiple ambisonics-binaural decoding solutions and report the best results for~\cite{morgadoNIPS18} in Table~\ref{tab:mono2binaural}.} Furthermore, manually inspecting the results, we find that the decoded binaural audio by~\cite{morgadoNIPS18} conveys spatial sensation, but it is less accurate and stable than our method. Our approach directly formulates the audio spatialization problem in terms of the two-channel binaural audio that listeners ultimately hear, which yields better accuracy. 

Our video results\footnote{\label{project_page}\url{http://vision.cs.utexas.edu/projects/2.5D_visual_sound/}}  show qualitative results including failure cases. Our system can fail when there are multiple objects of similar appearance, e.g. multiple human speakers. Our model incorrectly spatializes the audio, because the people are too visually similar. However, when there is only one human speaker amidst other sounds, it can successfully perform audio spatialization. Future work incorporating motion may benefit instance-level spatialization.

\begin{figure}
\centering
\begin{subfigure}{.5\textwidth}
  \centering
  \includegraphics[width=.95\linewidth]{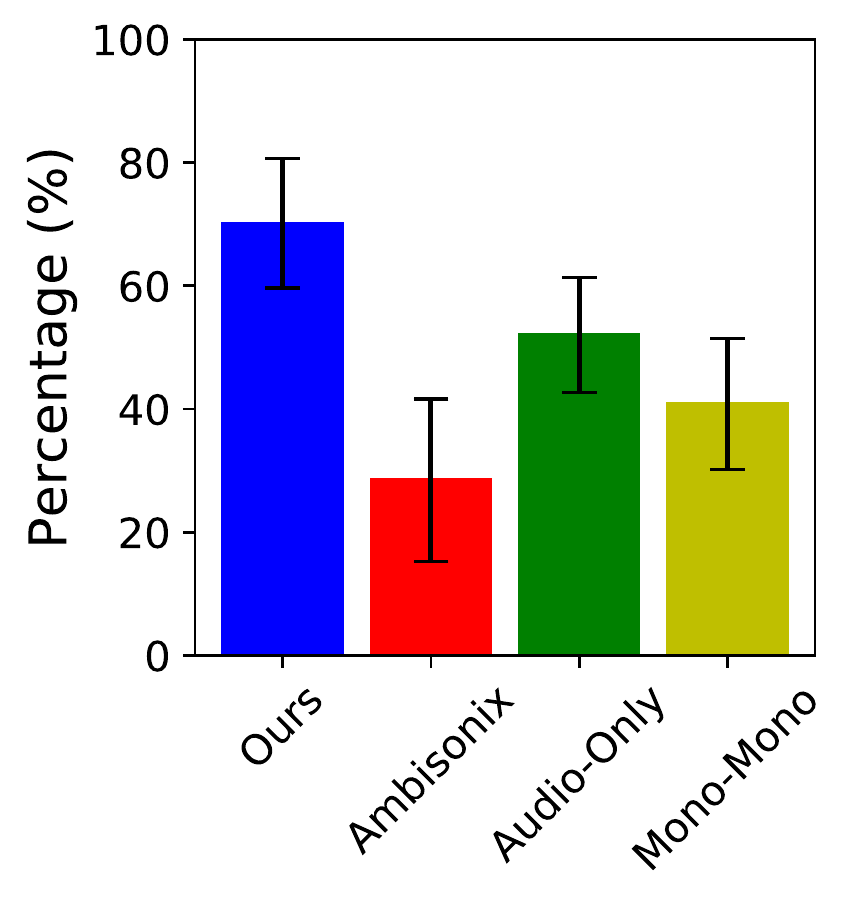}
  \vspace{-0.05in}
  \caption{User study 1}
  \label{fig:user_study_1}
\end{subfigure}%
\begin{subfigure}{.5\textwidth}
  \centering
  \includegraphics[width=.95\linewidth]{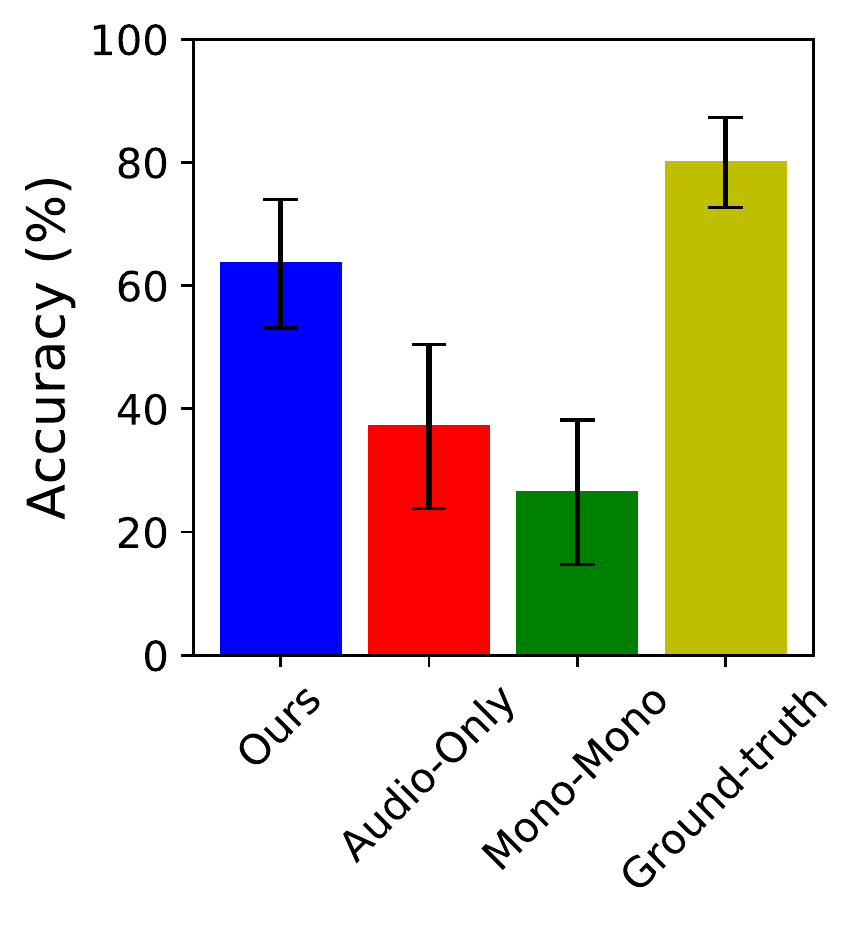}
  \vspace{-0.05in}
  \caption{User study 2}
  \label{fig:user_study_2}
\end{subfigure}
\caption{User studies to test how listeners perceive the predicted binaural audio.}
\label{fig:user_study}
\vspace{-0.1in}
\end{figure}

\begin{figure*}[t]
    \center
    \includegraphics[scale=0.38]{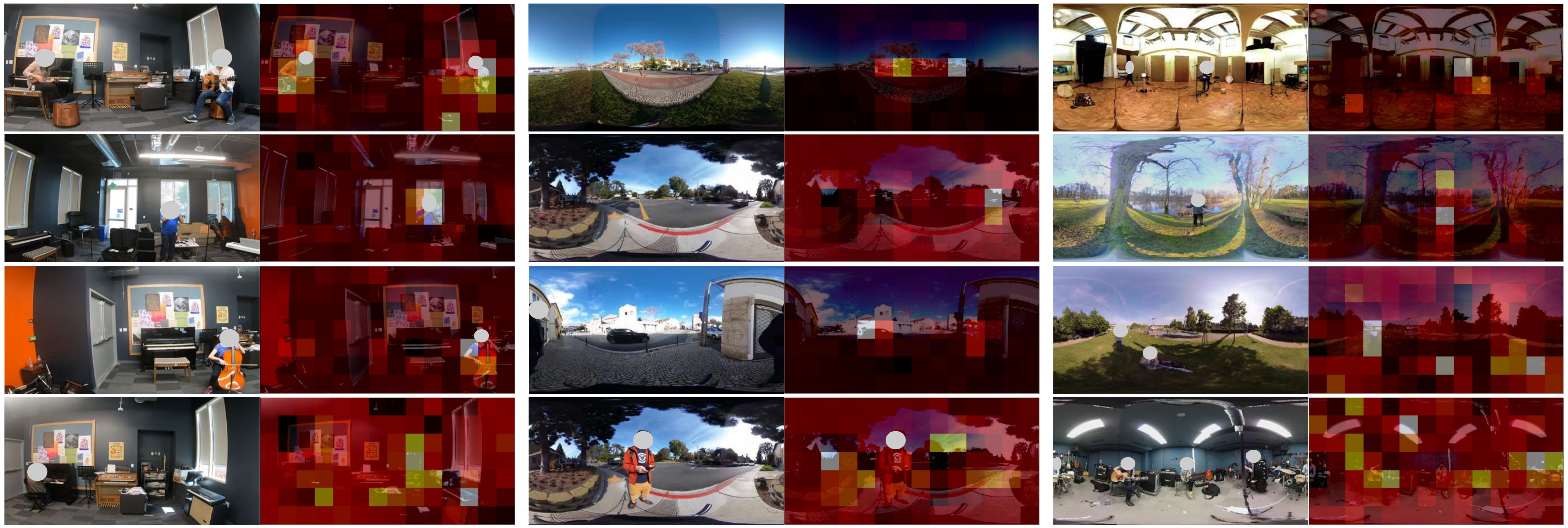}
    \caption{Visualizing the key regions the visual network focuses on when performing \textsc{mono2binaural} conversion. Each pair of images shows the frame accompanying the monaural audio (left) and the heatmap of the key regions overlaid (right).}   
    \label{fig:localization}
    \vspace{-0.1in}
\end{figure*}

\paragraph{User studies.}
Having quantified the advantage of our method in Table~\ref{tab:mono2binaural}, we now report real user studies. To test how well the predicted binaural audio makes a listener feel the 3D sensation, we conduct two user studies.

For the first study, the participants listen to a 10s ground-truth binaural audio and see the visual frame. Then they listen to two predicted binaural audios generated by our method and a baseline (Ambisonics, Audio-Only, or Mono-Mono). After listening to each pair, participants are asked which of the two creates a better 3D sensation that matches the ground-truth binaural audio. We recruited 18 participants with normal hearing. Each  listened to 45 pairs spanning all the datasets. Fig.~\ref{fig:user_study_1} shows the results. We report the percentage of times each method is chosen as the preferred one. We can see that the binaural audio generated by our method creates a more realistic 3D sensation.

For the  second user study, we ask participants to name the direction they hear a particular sound coming from. Using the FAIR-Play data, we randomly select 10 instrument video clips where some player is located in the left/center/right of the visual frames. We ask every participant to \emph{only listen} to the ground-truth or predicted binaural audio from our method or a baseline, and then choose the direction the sound of a specified instrument is coming from. Note that for this study, we input real mono audio recorded by the GoPro mic for binaural audio prediction. Fig.~\ref{fig:user_study_2} shows the results from the 18 participants. The true recorded binaural audio is of high quality, and the listeners can often easily perceive the correct direction. However, our predicted binaural audio also clearly conveys directionality. Compared to the baselines, ours presents listeners a much more accurate spatial audio experience.

\subsection{Localizing the Sound Sources}
\vspace*{-0.05in}
Does the network attend to the locations of the sound sources when performing binauralization? As a byproduct of our \textsc{mono2binaural} training, we can use the network to perform sound source localization. We use a mask of size $32 \times 32$ to replace image regions with image mean values, and forward the masked frame through the network to predict binaural audio. Then we compute the loss, and repeat by placing the mask at different locations of the frame. Finally, we highlight the regions which, when replaced, lead to the largest losses. They are considered the most important regions for \textsc{mono2binaural} conversion, and are expected to align with sound sources.

Fig.~\ref{fig:localization} shows examples. The highlighted key regions correlate quite well with sound sources. They are usually the instruments playing in the music room, the moving cars in street scenes, the place where an activity is going on, \etc. The final row shows some failure cases. The model can be confused when there are multiple similar instruments in view, or silent or noisy scenes. Sound sources in YT-Clean and YT-Music are especially difficult to spatialize and localize due to diverse and/or large number of sound sources.

\subsection{Audio-Visual Source Separation}
\vspace*{-0.05in}

Having demonstrated our predicted binaural audio creates a better 3D sensation, we now examine its impact on audio-visual source separation using the FAIR-Play dataset. The dataset contains object-level sounds of diverse sound making objects (instruments), which is well-suited for the Mix-and-Separate audio-visual source separation approach we adopt. We train on the held-out data of FAIR-Play, and test on 10 typical single-instrument video clips from the val/test set, with each representing one unique instrument in our dataset. We pairwise mix each video clip and perform separation, for a total of 45 test videos.

 \begin{table}
\centering
\vspace*{-0.1in}
\begin{tabular}{c?{0.5mm}c|c|c}
                     & SDR & SIR & SAR   \\ \specialrule{.12em}{.1em}{.1em}
Mono                 & 2.57   & 4.25 & 10.12\\ 
Mono-Mono            & 2.43  & 4.01 & 10.15\\ 
Predicted Binaural (Ours) & \textbf{3.01} & \textbf{5.03}  & \textbf{10.24} \\ \hline
GT Binaural (upper bound)     & 3.25   & 5.32  & 10.60 \\ 
\specialrule{.12em}{.1em}{.1em}
\end{tabular}
\caption{Audio-visual source separation results. SDR, SIR, SAR are reported in dB; higher is better.}
\label{tab:source_separation}
\vspace{-0.1in}
\end{table}

In addition to the ground truth binaural (upper bound) and the Mono-Mono baseline defined above, we compare to a \textbf{Mono} baseline that takes monaural audio as input and separates monaural audios for each source. Mono represents the current norm of performing audio-visual source separation using only single-channel audio~\cite{zhao2018sound,gao2018objectSounds,owens2018audio}. We stress that all other aspects of the networks are the same, so that any differences in performance can be attributed to our binauralization self-supervision. To evaluate source separation quality, we use the widely used mir eval library~\cite{raffel2014mir_eval}, and the standard metrics: Signal-to-Distortion Ratio (SDR), Signal-to-Interference Ratio (SIR), and Signal-to-Artifact Ratio (SAR). Table~\ref{tab:source_separation} shows the results. We obtain large gains by inferring binaural audio. The inferred binaural audio offers a more informative audio representation compared to the original monaural audio, leading to cleaner separation. See supp. video\textsuperscript{\ref{project_page}} for examples.

\vspace*{-0.05in}
\section{Conclusion}
\vspace*{-0.05in}

We presented an approach to convert single channel audio into binaural audio by leveraging object/scene configurations in the visual frames. The predicted 2.5D visual sound offers a more immersive audio experience. Our \textsc{mono2binaural} framework achieves state-of-the-art performance on audio spatialization. Moreover, using the predicted binaural audio as a better audio representation, we boost a modern model for audio-visual source separation. Generating binaural audio for off-the-shelf video can potentially close the gap between transporting audio and visual experiences, enabling new applications in VR/AR. As future work, we plan to explore ways to incorporate object localization and motion, and explicitly model scene sounds.

\vspace*{-0.15in}
\footnotesize
\paragraph{Acknowledgements:} Thanks to Tony Miller, Jacob Donley, Pablo Hoffmann, Vladimir Tourbabin, Vamsi Ithapu, Varun Nair, Abesh Thakur, Jaime Morales, Chetan Gupta from Facebook, Xinying Hao, Dongguang You, and the UT Austin vision group for helpful discussions.

{\small
\bibliographystyle{ieee}
\bibliography{ref_RG}
}

\normalsize
\newpage
\appendix
\noindent {\LARGE \textbf{Appendix}}

\section{Supplementary Video}
In our supplementary video\textsuperscript{\ref{project_page}}, we show (a) examples of our professional recorded binaural audios, (b) example results of binaural audio prediction, and (c) example results of audio-visual source separation.

\section{Details of \textsc{Mono2Binaural} Network}
Our \textsc{Mono2Binaural} network consists of a visual branch and an audio branch. The visual branch takes images of dimension 224x448x3 as input to extract a feature map of dimension 14x7x512 through ImageNet pre-trained ResNet-18 network. The visual feature map is then passed though a 1x1 convolution layer to reduce the channel dimension, producing a feature map of dimension 14x7x8. 

The audio branch is of a U-NET style architecture, namely an encoder-decoder network with skip connections. It consists of 5 convolution layers and 5 up-convolution layers. All convolutions and up-convolutions use 4 x 4 spatial filters applied with stride 2, and followed by a BatchNorm layer and a ReLU. After the last layer in the decoder, an up-convolution is followed by a Sigmoid layer to bound the values of the complex mask. The encoder uses leaky ReLUs with a slope of 0.2, while ReLUs in the decoder are not leaky. Skip connections are added between each layer $i$ in the encoder and layer $n - i$ in the decoder, where $n$ is the total number of layers. The skip connections concatenate activations from layer $i$ to layer $n - i$.

\section{Details of \textsc{Mix-and-Separate} Network}
For audio-visual source separation, we use the same base architecture as our \textsc{mono2binaural} network except that now the input to the network is a pair of training video clips. Two visual branches of shared weights are used and each takes the frame of one video as input to extract visual features. The audio branch takes the mixed audio as input to extract audio features, and is combined with the visual features to predict a mask for each video. Following~\cite{zhao2018sound}, we use ratio masks and log magnitude spectrograms. For ratio masks, the ground truth mask of a video is calculated as the ratio of the magnitude spectrogram of the target sound and the mixed sound.

\section{Implementation Details}
For \textsc{mono2binaural} training, we randomly sample audio segments of length 0.63s from each 10s audio clip and normalize each segment's RMS level to a constant value. Then we obtain a complex spectrogram of size $257 \times 64 \times 2$ for each channel. For each sampled audio segment, the center video frame is used as the accompanying visual frame and resized to $480 \times 240$. We randomly crop $448 \times 224$ images and use color and intensity jittering as data augmentation. The network is trained using an Adam~\cite{adamsolver} optimizer with weight decay $5 \times 10^{-4}$ and batch size 256. The starting learning rate is set to 0.001, and decreased by 6\% every 10 epochs and trained for 1,000 epochs. We use smaller starting learning rate 0.0001 for ResNet-18 because it is pre-trained on ImageNet.

For audio-visual source separation training, we randomly sample pairs of video and take an audio segment of length 2.55s from each video. We mix the two audio segments, and obtain a log magnitude spectrogram of size $257 \times 256$ for each channel. A random frame within each audio segment is used as the accompanying visual frame for both videos. The network is trained with batch size 128 and learning rate 0.001, and a smaller learning rate 0.0001 is used for ResNet-18.

\begin{figure}
\centering
\begin{subfigure}{\textwidth}
  \centering
  \includegraphics[width=.93\linewidth]{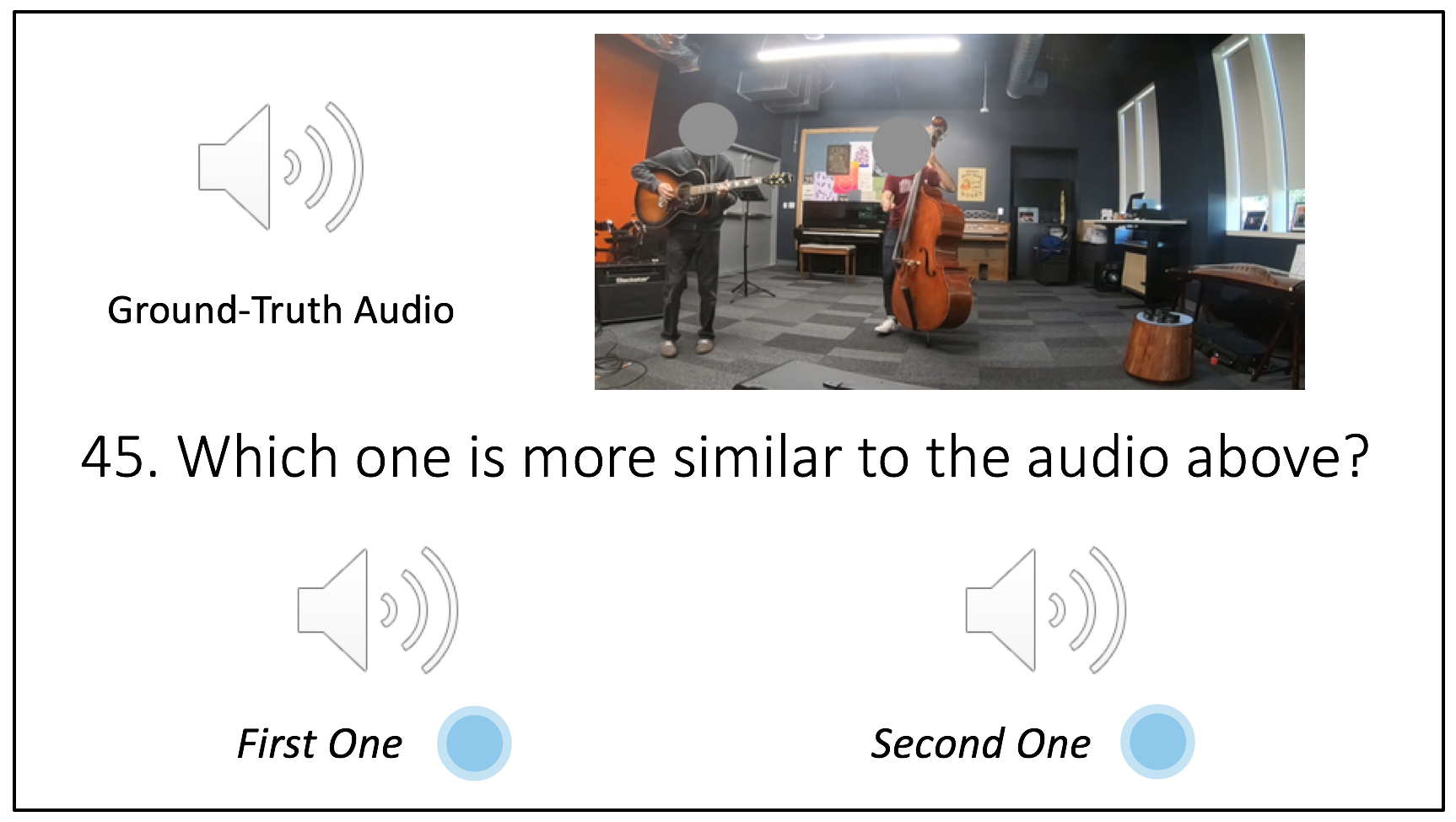}
  \caption{Example of user study 1}
  \label{fig:user_study_1_interface}
\end{subfigure}%
\hfill
\vspace{0.1in}
\begin{subfigure}{\textwidth}
  \centering
  \includegraphics[width=.93\linewidth]{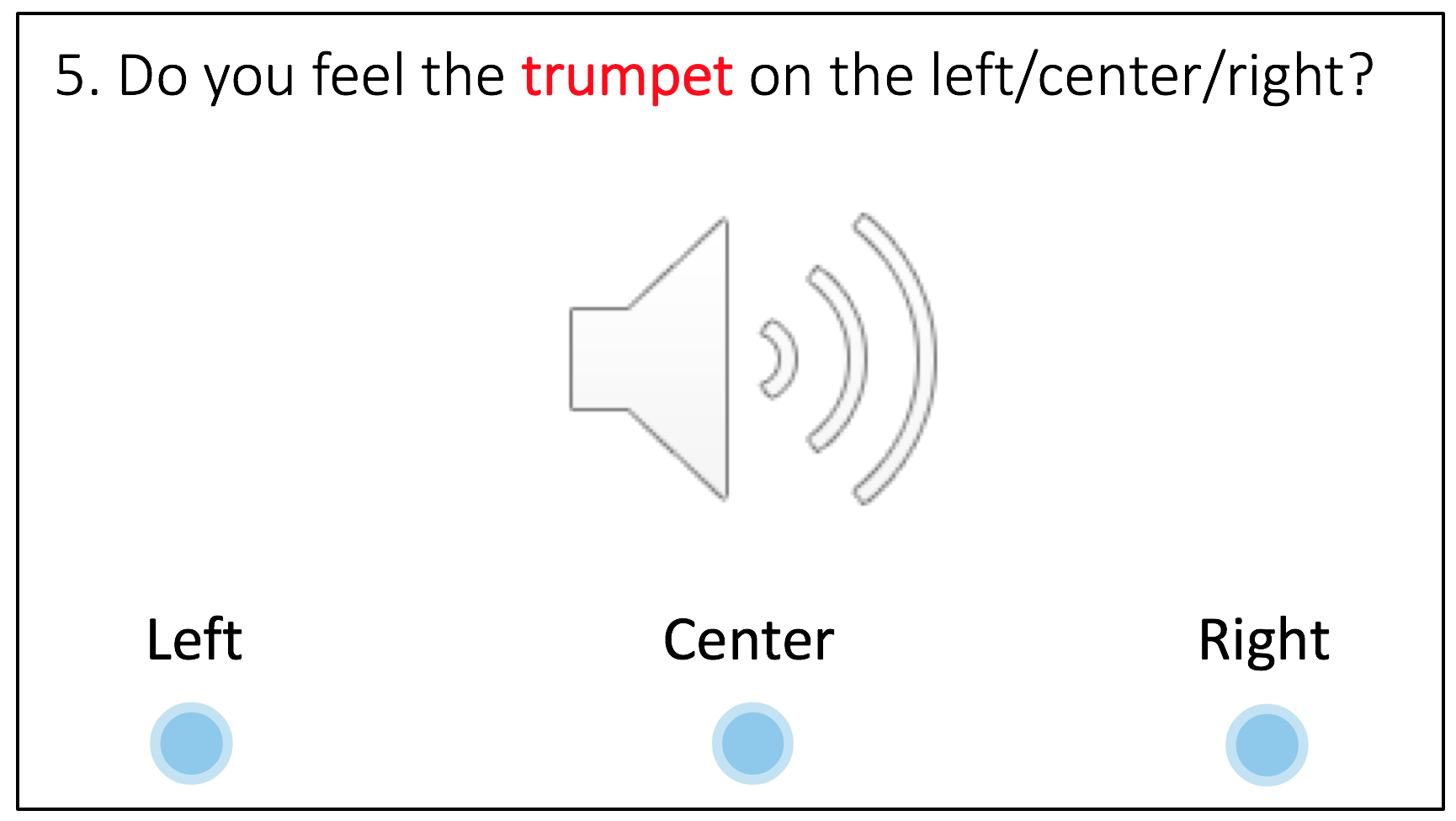}
  \caption{Example of user study 2}
  \label{fig:user_study_2_interface}
\end{subfigure}
\caption{Examples of the interface for the two user studies to test how listeners perceive the predicted binaural audio.}
\label{fig:user_study_interface}
\end{figure}

\section{User Study Interface}
In Fig.~\ref{fig:user_study_interface}, we show examples of our user study interface. In the first user study (Fig.~\ref{fig:user_study_1_interface}), the participants listen to a 10s ground-truth binaural audio and see the accompanying visual frame. Then they listen to two predicted binaural audios generated by our method or a baseline (Ambisonics, Audio-Only, or Mono-Mono). After listening to each pair, participants are asked whether the first one or the second one creates a better 3D sensation that matches the ground-truth binaural audio. In the second user study, we ask every participant to \emph{only listen} to the ground-truth or predicted binaural audio from our method or a baseline, and then choose the direction the sound of a specified instrument is coming from. For example, as shown in Fig.~\ref{fig:user_study_2_interface}, participants first listen to the audio, and then they are asked to choose whether they feel the trumpet on the left, in the center, or on the right.

\end{document}